\documentclass[11pt,a4paper]{article}
\usepackage{times,latexsym}
\usepackage{url}
\usepackage[T1]{fontenc}
\usepackage[utf8]{inputenc}

\usepackage[acceptedWithA]{tacl2021v1}

\usepackage{booktabs}
\usepackage{array}
\usepackage{xcolor}
\usepackage{graphicx}
\usepackage{placeins}
\usepackage{float}
\makeatletter
\renewcommand{\confidential}{\normalcolor\confidentialtext}
\let\blimpLowish\AtPageLowishCenter
\renewcommand{\AtPageLowishCenter}[1]{\blimpLowish{\normalcolor#1}}
\makeatother

\title{BLiMP-RO: A Benchmark of Linguistic Minimal Pairs for Romanian}

\author{
    \textbf{Radu-Stefan Popa}\textsuperscript{1},
    \textbf{Valentin Feroiu}\textsuperscript{1},
    \textbf{Adina Camelia Bleotu}\textsuperscript{1,2},
    \textbf{Sergiu Nisioi}\textsuperscript{1}
    \\
    \\
    \textsuperscript{1}University of Bucharest \\
    \textsuperscript{2}Universit\"at Wien
}

\date{}

\begin{document}
\maketitle

\begin{abstract}
Evaluating grammatical knowledge in language models requires benchmarks that cover diverse linguistic phenomena and are validated against human judgments. We introduce BLiMP-RO, a Romanian minimal-pair benchmark covering 50 grammatical phenomena across ten domains. It comprises 493 linguist-constructed sentence pairs, evaluated by native speakers in five experiments, and 5,942 additional semi-automatically generated, human-validated pairs. We further investigate grammatical generalization under limited linguistic input by training GPT-2 and GPT-BERT models from scratch on Romanian BabyBabelLM data and a newly collected corpus of 10 million words from child-targeted video captions. Increasing caption training data yields approximately 9 percentage points of accuracy improvement per tenfold increase in corpus size. At matched data sizes, BabyBabelLM generally outperforms video captions, while GPT-BERT consistently outperforms GPT-2. Our findings highlight how training data quantity, composition, and model architecture shape grammatical generalization in Romanian.
\end{abstract}

\section{Introduction}

Minimal-pair benchmarks test whether a language model assigns a higher probability to a grammatical sentence than to a minimally different ungrammatical one \citep{marvin-linzen-2018,warstadt-etal-2020}. They build on acceptability judgments, a standard source of evidence in syntax \citep{chomsky-1965,schutze-2016}, and probe agreement and syntactic dependencies that overall predictive performance does not isolate \citep{linzen-etal-2016,wilcox-etal-2018,wilcox-etal-2024}. Minimal pairs can be used to examine the point at which grammatical knowledge emerges when language models are trained on smaller, developmentally-motivated corpora. As such, they must have broad linguistic coverage and evidence that speakers recognize the intended contrasts.

For Romanian, existing minimal-pair evaluation is limited. MultiBLiMP \citep{jumelet-etal-2026-multiblimp} contains 2,056 Romanian pairs derived from treebanks, but they focus on agreement involving subjects, verbs, and participles and have not been validated with native-speaker judgments.
The dataset, however, does not cover typical properties of Romanian grammar such as differential object marking (DOM), clitics and clitic doubling, strict negative concord, referential null subjects, or subjunctive complements in contexts where other Romance languages prefer an infinitive \citep{panadindelegan2013,dobroviesorin1994}.

We introduce BLiMP-RO, a linguist-constructed benchmark covering 50 phenomena across ten grammatical domains (see \autoref{tab:inventory}). We develop two datasets: an initial hand-crafted set of 10 pairs of sentences per phenomenon (500 items) and an extended set constructed semi-automatically consisting of 120 pairs of sentences per phenomenon (6000 items). The initial set is used for acceptability rating with at least 50 human participants for each item across five experiments. After further validation, we retain 493 pairs. Human accuracy (96\%) and agreement (93.1\%) are computed on this dataset. The extended set is constructed using few shot prompting using Claude Opus 5.5 and has been validated by 3 human annotators, leaving a total of 5,942 valid sentence pairs. The complete set is mainly used to compare language models.

We use BLiMP-RO to evaluate two pretrained Romanian models and two pretrained multilingual models. We additionally study grammatical generalization under limited-input pretraining. Following the BabyLM approach of training relatively small models on developmentally motivated corpora \citep{warstadt-etal-2023-findings}, we train Romanian GPT-2 \citep{radford-etal-2019-language} and GPT-BERT models from scratch using two data sources. The first is the Romanian portion of BabyBabelLM \citep{jumelet-etal-2026-babybabellm}, which combines child-directed speech and children's books with subtitle-based fallback text. The second is a curated corpus of 10 million Romanian words from captions of child-targeted videos discovered through YouTube Kids profiles for ages 0--4 and 5--8, followed by collection from eligible channels.

Our contributions can be summarized as follows: 1. the creation of \textbf{BLiMP-RO}\footnote{BLiMP-RO, the human judgments and the code will be released under CC-BY-4.0.} a Romanian grammatical benchmark that covers typical syntactic and morphosyntactic phenomena missing from existing Romanian minimal-pair evaluations. Human validation obtains 96\% accuracy in categorical judgments and agrees with intended preference.
2.  We assemble a Romanian child-targeted caption corpus of 10 million words from children's content captions (about 2,100 hours) and train BabyLM-style GPT-2 and GPT-BERT models on both these captions and BabyBabelLM data. 
3. We compare pretrained Romanian models with models trained from scratch, examine their correspondence with human judgments, and analyze how grammatical accuracy changes with the amount and source of training data. Our results confirm that books are a more adequate corpus for grammar acquisition in baby language models.



\section{Related Work}

Besides English, minimal-pair benchmarks exist for Chinese \citep[CLiMP and SLING;][]{xiang-etal-2021,song-etal-2022}, Japanese \citep[JBLiMP;][]{someya-oseki-2023}, Russian \citep[RuBLiMP;][]{taktasheva-etal-2024}, Dutch \citep[BLiMP-NL;][]{suijkerbuijk-etal-2025}, Turkish \citep[TurBLiMP;][]{basar-etal-2025-turblimp}, Irish \citep[Irish-BLiMP;][]{mcgiff-etal-2026-irish} and Italian \citep{barbini-etal-2025,bressan-etal-2026}. While all rely on minimal pairs, these benchmarks differ first in how they build pairs. The original English BLiMP uses expert templates and a feature-annotated vocabulary for 67 paradigms of 1,000 pairs \citep{warstadt-etal-2020}. BLiMP-NL combines ten hand-written pairs per paradigm with 90 semi-automatically generated, manually validated pairs \citep{suijkerbuijk-etal-2025}. SLING and RuBLiMP transform treebank or corpus sentences. MultiBLiMP derives subject agreement pairs for 101 languages from Universal Dependencies and UniMorph. JBLiMP takes pairs from the theoretical literature, and Irish-BLiMP constructs them manually. BLiMP-RO has about ten hand-written pairs per construction. 

Benchmarks also differ in how they check pairs with people. BLiMP \citep{warstadt-etal-2020} collected 20 judgments for each of five pairs per paradigm; a majority agreed with the label for 96.4\% of the pairs, and individual raters in 88.6\% of their judgments. ZhoBLiMP \citep{liu-etal-2026-systematic} collected five forced-choice judgments for each of five pairs per paradigm. JBLiMP \citep{someya-oseki-2023} removed candidates that failed majority validation, and RuBLiMP \citep{taktasheva-etal-2024} asked linguistics students whether a pair isolates its target contrast. BLiMP-NL \citep{suijkerbuijk-etal-2025} and TurBLiMP \citep{basar-etal-2025-turblimp} collected graded acceptability ratings and kept agreement with the label apart from measured acceptability. For BLiMP-RO, 50 to 56 native speakers judged every item in a forced choice.

Regarding model evaluation, human validation establishes whether benchmark items reflect the intended contrasts, but model comparisons also depend on how sentences are scored. Length and frequency adjustments can change minimal-pair results \citep{lau-etal-2017,liu-etal-2026-systematic,tjuatja-etal-2025-goes}. Summed log probability penalizes longer sentences. The syntactic log-odds ratio, SLOR, subtracts a unigram log probability and divides by length \citep{pauls-klein-2012-large}. \citet{lau-etal-2017} compared several scores computed over the whole sentence. Of these, SLOR best predicted the ratings that people gave, and \citet{suijkerbuijk-etal-2025} recommends it for comparing models. Tokenwise pseudo-log-likelihood for masked models \citep{shin-etal-2019-effective,salazar-etal-2020-masked} inflates the scores of words split into several tokens, which a word-level variant avoids \citep{kauf-ivanova-2023-better}, and model-human alignment depends on how the two are compared \citep{hu-etal-2024-language}. Following work on models trained with child-directed input \citep{huebner-etal-2021-babyberta,warstadt-etal-2023-findings}, we also train models on children's video captions and the Romanian part of BabyBabelLM \citep{jumelet-etal-2026-babybabellm}.

\begin{table*}[!htb]
\centering
\newcommand{\dom}[2]{\noindent\textbf{#1}\hfill\mbox{#2}\par\nopagebreak\vspace{1pt}}
\newcommand{\ph}[1]{\hangindent=1em\hangafter=1\noindent #1\par}
\newcommand{\gap}{\par\vspace{6pt}}
\begin{tabular}{@{}p{5.6cm}@{\hspace{0.4cm}}p{5.3cm}@{\hspace{0.4cm}}p{4.3cm}@{}}
\toprule
\begin{minipage}[t]{\linewidth}\raggedright
\dom{Morphosyntactic agreement}{77}
\ph{subject-verb}
\ph{demonstrative}
\ph{adjective}
\ph{participial agreement}
\ph{adjectival article \emph{cel}}
\ph{anaphor agreement (emphatic reflexive \emph{însuși})}
\ph{doubled object and clitic}
\ph{ordinal numeral}
\gap
\dom{Differential object marking}{30}
\ph{\emph{pe} in object relatives}
\ph{\emph{pe} and clitic across referential noun types}
\ph{object marking across animacy classes}
\gap
\dom{Clitics}{46}
\ph{clitic doubling}
\ph{resumptive clitics in relatives and \emph{pe~cine} questions}
\ph{clitic order and the Person Case Constraint}
\ph{clitics with modal verbs}
\ph{clitics in negative imperatives}
\end{minipage}
&
\begin{minipage}[t]{\linewidth}\raggedright
\dom{Binding theory}{30}
\ph{case condition (gerund against finite verb after a reflexive)}
\ph{C-command condition (antecedent of an emphatic reflexive)}
\ph{locality of reflexives}
\gap
\dom{Argument structure}{29}
\ph{transitivity}
\ph{passivizability}
\ph{inherent reflexive verbs}
\gap
\dom{Word order and movement}{131}
\ph{relativization (prepositions in relatives)}
\ph{demonstrative placement}
\ph{topicalization (doubled preverbal objects)}
\ph{definite article placement}
\ph{indefinite article placement}
\ph{preverbal depictives}
\ph{auxiliary and verb order}
\ph{adpositional phrases (nominal dependents)}
\ph{extraction from complex noun phrases}
\ph{extraction from adjuncts}
\ph{extraction from coordinations}
\ph{extraction from embedded questions}
\ph{extraction from clausal subjects}
\end{minipage}
&
\begin{minipage}[t]{\linewidth}\raggedright
\dom{Mood}{50}
\ph{subjunctive-selecting predicates}
\ph{indicative-selecting predicates}
\ph{subjunctive and infinitive}
\ph{conditional}
\ph{future}
\gap
\dom{Negation}{29}
\ph{negative concord}
\ph{negative additive \emph{nici}}
\ph{negative imperatives}
\gap
\dom{Nominal morphology}{61}
\ph{gender and number}
\ph{plural forms}
\ph{nominalization}
\ph{genitive and dative}
\ph{cardinal numerals with \emph{de}}
\ph{vocative}
\gap
\dom{Null subjects}{10}
\ph{null against overt subject}
\end{minipage}
\\
\bottomrule
\end{tabular}
\caption{Domains and phenomena/subdomains of BLiMP-RO. The number after a domain counts its distinct pairs.}
\label{tab:inventory}
\end{table*}

\section{BLiMP-RO}

\subsection{Phenomena}

We selected the phenomena from our knowledge of Romanian and the Grammar of the Romanian
Language \citep{galr2008set}. We started from English BLiMP and
Dutch BLiMP-NL \citep{warstadt-etal-2020,suijkerbuijk-etal-2025} and adapted the inventory to Romanian morphology and syntax. 
Table~\ref{tab:inventory} groups the 50 phenomena into ten domains by the principal grammatical contrast of their items.

Like other \textit{BLiMPs}, BLiMP-RO tests agreement, argument structure, nominal morphology and extraction, here from complex noun phrases, adjuncts, coordinations, embedded questions and clausal subjects \citep{ross-1967-constraints}. In two factorial experiments, \citet{schoenmakersstoica2024} found island effects in Romanian for adjuncts, embedded questions with \emph{dacă} `if', complex noun phrases and subjects. BLiMP-RO also tests phenomena that English BLiMP does not cover. Romanian marks some direct objects with \emph{pe}, often together with clitic doubling, depending on the referential type of the nominal, animacy, specificity and discourse prominence \citep{aissen-2003,cornilescu2000prepositional,tigau2010account,tigau2011syntax,irimia2020structural,hillmardale2021,LopezOteroBleotu2025}. Its pronominal clitics have restricted positions and combinations, including the person restrictions in clitic clusters known as the Person Case Constraint \citep{bonet-1991-morphology,dobroviesorin1994,nevinssavescu2010}. BLiMP-RO also tests the competition between infinitives and subjunctives. As in other languages of the Balkan Sprachbund, \emph{să}-subjunctives appear in many contexts where other Romance languages use an infinitive \citep{BilbiieMardale2018,cotfas2016syntax,cotfas2016subjunctive}, but infinitives remain available under restricted lexical, syntactic and register conditions.

Romanian is a strict negative concord language. In finite clauses, negative words such as \emph{nimeni} `nobody' co-occur with preverbal \emph{nu} and express a single negation \citep{giannakidouzeijlstra2017landscape,falausnicolae2016}, as in \emph{Nimeni nu a venit} `Nobody came', literally `nobody not has come'. Romanian is also a pro-drop language. It permits referential null subjects, and an overt subject pronoun can mark contrast, emphasis or a change of reference \citep{panadindelegan2013,dobrogi2013,istrate2024subject,frascarelli2021,bleotu2026contrast}. Our null-subject items compare an overt pronoun with its omission in a discourse context.

\subsection{Categorical and Preference Pairs}
\label{sec:types}

Most items oppose a sentence intended as grammatical to one intended as ungrammatical, and we call them \emph{categorical}. In \emph{preference} items, both sentences are intended as grammatical and one is expected to be strongly preferred as more natural. They contrast the subjunctive against the infinitive and a null against an overt subject. The answer key designates one sentence of each item as the target, intended as grammatical or preferred, and the other as the alternative. We assign the type by phenomenon.
This gives 20 preference items and 488 categorical items. 

\section{Acceptability Ratings}

\subsection{Procedure and Participants}

We created about ten pairs per phenomenon, aiming to keep the lexical and structural material of the two sentences constant while varying the target property.

We ran five forced-choice experiments in PCIbex
\citep{zehr-schwarz-2018-penncontroller}. After giving informed consent and completing a short questionnaire on their linguistic background, participants were instructed to choose the more natural variant and completed two practice trials. Each trial showed the question \emph{Care variantă este mai naturală?} `Which variant is more natural?' with the two sentences as options. Within a session, participants saw every item of the experiment once, in a shuffled order. 
The order of the two sentences of an item was fixed, with the target first in half of the trials.

The study was open to native speakers of Romanian aged 18 or over, recruited among university students for a bonus in a language or linguistics course. In total, we conducted five experiments, each with a set of 56, 53, 50, 51, and 50 disjoint participants, totaling 260 different people. In the background questionnaire, 256 sessions report Romanian as their first language, the median of the 258 reported ages is 20, ranging from 18 to 56, and 206 sessions report female gender, 51 male and 3 another. We kept all sessions. Excluding the six sessions with accuracy below 80\% would raise mean categorical item accuracy from 96\% to 96.7\%.

The participants across all five experiments saw 528 items, 100 in each of Experiments 1 to 3, 120 in Experiment~4 and 108 in Experiment~5. We analyze 508 pairs\footnote{An item is a pair as presented in one experiment. Fifteen pairs were presented in two experiments, so the 508 items represent 493 distinct pairs.}, leaving out 13 items later judged to contain flaws, such as a second agreement error in one sentence, and eight items without a consistent key, with one item in both groups. Five nominalization pairs of Experiment 1 were repeated in Experiment 5 and ten clitic-doubling pairs of Experiment 2 in Experiment 3. Sentences have 2 to 13 words, with a median of 6.


\subsection{Agreement with the Target}
On categorical items, participants chose the target in 24,336 of 25,348 trials, or 96\%. When items are weighted equally, we obtain a mean human item accuracy of 96\% [95.1, 96.8] on the categorical items, with the bootstrap intervals of Section~\ref{sec:models} \cite{dror-etal-2018-hitchhikers}. A majority chose the target in all 20 preference items and in 487 of the 488 categorical items. The exception is a DOM item across animacy classes, whose alternative the linguistic annotation marks as degraded rather than ungrammatical. On the 20 preference items, we obtain a mean agreement with the target of 97.1\% [94.8, 98.9]. Agreement with the target on the ten clitic-doubling pairs repeated in Experiments 2 and 3 is 97.2\% and 93.0\%, with Pearson $r = 0.61$ across pairs, and on the five nominalization pairs repeated in Experiments 1 and 5, 93.9\% and 95.2\%, with $r = 0.94$.

\subsection{Extended Set}
\label{sec:extended}
\begin{figure*}[t]
\centering
\includegraphics[width=\textwidth]{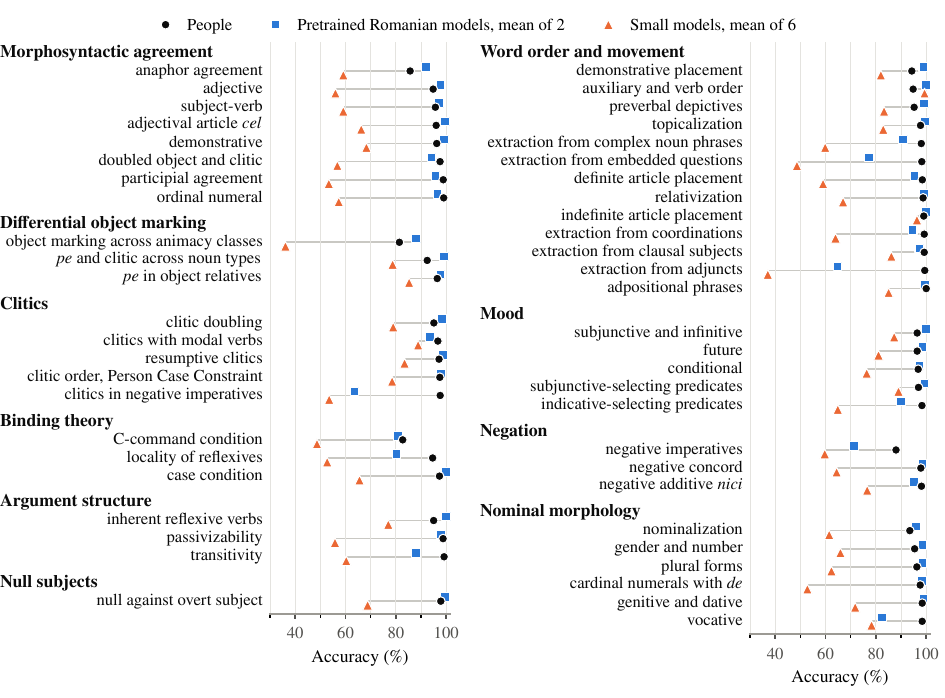}
\caption{Accuracy (\%) on each phenomenon, grouped by the domains of Table~\ref{tab:domains}.
Circles are people, squares the mean of the two pretrained Romanian models and triangles the mean of
GPT-2 and GPT-BERT trained on captions 1M, BBL 1M and captions 10M (Table~\ref{tab:main}). People are on the 508
items, models on all pairs, with the 5,942 pairs that humans accepted. A gray line joins the three marks of a phenomenon. Within a
domain, rows go from the lowest human accuracy at the top to the highest at the bottom. Names are shortened from
Table~\ref{tab:inventory}; anaphor agreement and the C-command condition test emphatic reflexives. The
axis starts at 30. Table~\ref{tab:phenomena} gives the values of each
model.}
\label{fig:phenomena}
\end{figure*}

Besides the 493 pairs, we have 6000 further pairs for the same 50 phenomena. Between two and three human annotators judged the extended data pairs on a custom web platform created for this experiment. For each pair, a person indicated whether the target sentence is correct, both sentences are, or neither is. When both sentences are correct, the preferred variant was required to be marked. Two persons who judged the same pair agreed on whether to accept it in 97.4\% of cases. We kept only the 5,942 pairs that had perfect agreement. Tables~\ref{tab:domains},\ref{tab:selections}, \ref{tab:sources}
and~\ref{tab:phenomena} and Figures~\ref{fig:phenomena} and~\ref{fig:curve} report models on all 6,435 pairs, and Table~\ref{tab:main} reports the 508 items and the 5,942 pairs in separate columns.

\begin{table}[!t]
\centering
\setlength{\tabcolsep}{4pt}
\begin{tabular}{@{}l r@{\ }l c r r@{}}
\toprule
& \multicolumn{4}{c}{BLiMP-RO} & Multi \\
\cmidrule(lr){2-5}
& \multicolumn{2}{c}{Init.} & Agr. & Ext. & \\
\midrule
Humans & 96 & [95.1,\,96.9] & 93.1 &  & \\
\addlinespace[2pt]
\multicolumn{6}{@{}l}{\emph{GPT-2}} \\
\hspace{0.5em}Cap. 1M & 65 & [60.8,\,68.6] & 63.7 & 62 & 59 \\
\hspace{0.5em}BBL 1M & 65 & [62.5,\,68.1] & 64.8 & 68 & 64 \\
\hspace{0.5em}Cap. 0--4 & 71 & [67.5,\,74.5] & 69.7 & 70 & 66 \\
\hspace{0.5em}Cap. 5--8 & 69 & [65.0,\,71.8] & 67.5 & 68 & 64 \\
\hspace{0.5em}Cap. 10M & 73 & [69.5,\,75.5] & 71.2 & 72 & 68 \\
\addlinespace[2pt]
\multicolumn{6}{@{}l}{\emph{GPT-BERT}} \\
\hspace{0.5em}Cap. 1M & 70 & [66.5,\,73.6] & 68.8 & 67 & 62 \\
\hspace{0.5em}BBL 1M & 71 & [68.2,\,74.0] & 70.4 & 71 & 66 \\
\hspace{0.5em}Cap. 0--4 & 75 & [72.3,\,78.2] & 73.9 & 72 & 68 \\
\hspace{0.5em}Cap. 5--8 & 76 & [72.5,\,78.5] & 74.3 & 72 & 67 \\
\hspace{0.5em}Cap. 10M & 77 & [74.6,\,80.3] & 76.0 & 74 & 71 \\
\addlinespace[2pt]
\multicolumn{6}{@{}l}{\emph{Word trigrams}} \\
\hspace{0.5em}Cap. 1M & 55 & [51.3,\,59.0] & 55.5 & 54 & 51 \\
\hspace{0.5em}BBL 1M & 62 & [58.2,\,65.8] & 61.9 & 59 & 60 \\
\hspace{0.5em}Cap. 0--4 & 59 & [55.6,\,63.2] & 59.6 & 57 & 55 \\
\hspace{0.5em}Cap. 5--8 & 59 & [55.0,\,62.7] & 58.5 & 57 & 54 \\
\hspace{0.5em}Cap. 10M & 61 & [57.6,\,65.3] & 61.2 & 58 & 58 \\
\midrule
RoGPT2 & 90 & [88.2,\,92.5] & 88.1 & 94 & 95 \\
RoBERT & 92 & [90.4,\,94.1] & 89.8 & 95 & 98 \\
\midrule
EuroLLM & 82 & [79.6,\,84.9] & 80.6 & 91 & 96 \\
Salamandra & 82 & [79.0,\,84.8] & 80.2 & 90 & 94 \\
\bottomrule
\end{tabular}
\caption{SLOR Accuracy (\%) for different models and pre-training scenarios. The column \emph{Init.} reports accuracy on the 508 items of BLiMP-RO, with 95\% bootstrap
intervals over items and runs. \emph{Agr.}: agreement with people, how often a model chose the same sentence as a
participant, over the 26,358 answers; for people, how often two participants chose the same sentence on an item, over
671,868 pairs. \emph{Ext.}: accuracy on the extended set of
5,942 further pairs that participants accepted in a validation study. \emph{Multi}: accuracy on MultiBLiMP. Trained models are averaged over five runs. Cap.\ is captions and BBL is BabyBabelLM.}
\label{tab:main}
\end{table}

\section{Models and Training Data}
\label{sec:models}

We evaluate two pretrained Romanian transformers, RoGPT2
\citep{niculescu-etal-2021-rogpt2}, a \mbox{GPT-2} with 124M parameters, 
and RoBERT \citep{masala-etal-2020-robert}, and two larger multilingual models that include Romanian, EuroLLM-9B in its 2512 release \citep{martins-etal-2025-eurollm9b,ramos-etal-2026-eurollm22b} and Salamandra-7B \citep{gonzalez-agirre-etal-2025-salamandra}.
Additionally, we train a trigram language model using Kneser--Ney smoothing \citep{kneser-ney-1995-improved,chen-goodman-1999-empirical} as a baseline, and two neural models, GPT-2 and GPT-BERT, from scratch on two sources. The first source is 10 million words of creator-provided and automatically generated Romanian captions from 5,356 videos (2,094 hours) from YouTube Kids.
They run at 80 words per minute of video. We found the videos through category pages, recommendations, fixed searches and channel pages under the YouTube Kids profiles for ages 0--4 and 5--8, and then collected the other eligible videos of the channels found this way. We first collected the videos of the 0--4 profile and then those of the 5--8 profile, skipping any video already collected for ages 0--4, since the YouTube Kids setting for ages 5--8 also includes the content for ages 0--4. 
We kept only Romanian captions written by the creators and Romanian automatic captions, excluding translated captions. We checked the language of each caption track with the lingua detector \citep{stahl2025lingua}. Automatic captions make up 95.2\% of the videos and 97.5\% of the words. All videos have Romanian speech. Many programs were first made in another language and dubbed into Romanian by Romanian voice actors; they account for 82.3\% of the words, programs made in Romanian for 4.9\%, and channels that publish both for 12.8\%. The largest channel provides 13.3\% of the words for ages 0--4 and 20.0\% of the words for ages 5--8. To estimate caption accuracy, a native speaker transcribed 40 one-minute windows of the videos from the audio. Against these transcriptions, the captions have a word error rate of 4.0\% [2.9, 5.5], and 9 of the 864 words that carry agreement (1.0\%) change their person, number or gender (Appendix~\ref{app:details}). 

\begin{table*}[!htb]
\centering
\setlength{\tabcolsep}{2.5pt}
\begin{tabular}{@{}l rr rrr rrr rr@{}}
\toprule
& & & \multicolumn{3}{c}{GPT-2} & \multicolumn{3}{c}{GPT-BERT} & \multicolumn{2}{c}{Pretrained} \\
\cmidrule(lr){4-6}\cmidrule(lr){7-9}\cmidrule(l){10-11}
& Humans & Pairs & Cap.\,1M & BBL & Cap.\,10M & Cap.\,1M & BBL & Cap.\,10M & RoGPT2 & RoBERT \\
\midrule
Agreement & 95.4 & 1011 & 54.7 & 55.4 & 65.2 & 58.7 & 55.7 & 67.5 & 95.5 & 97.6 \\
DOM & 90.1 & 389 & 55.2 & 65.8 & 67.1 & 68.0 & 70.6 & 73.3 & 94.1 & 95.9 \\
Clitics & 96.5 & 727 & 68.9 & 76.8 & 77.6 & 77.2 & 83.1 & 79.2 & 92.9 & 90.6 \\
Binding & 91.5 & 369 & 53.7 & 59.0 & 58.5 & 49.7 & 55.6 & 59.6 & 88.6 & 86.2 \\
Arguments & 97.6 & 383 & 56.1 & 62.5 & 68.0 & 62.9 & 67.5 & 70.0 & 92.2 & 98.4 \\
Word order & 97.9 & 1657 & 67.4 & 72.3 & 76.2 & 69.8 & 77.3 & 76.5 & 91.9 & 95.4 \\
Mood & 97.0 & 649 & 75.7 & 78.9 & 82.6 & 79.9 & 79.8 & 81.4 & 96.8 & 97.2 \\
Negation & 94.9 & 363 & 53.9 & 66.4 & 71.7 & 65.5 & 70.1 & 75.6 & 88.2 & 90.4 \\
Nominal & 96.3 & 765 & 58.4 & 66.6 & 69.1 & 58.8 & 66.8 & 72.4 & 96.5 & 94.5 \\
Null subjects & 97.8 & 122 & 57.7 & 47.0 & 74.4 & 79.8 & 69.3 & 84.3 & 99.2 & 100.0 \\
\bottomrule
\end{tabular}
\caption{Accuracy (\%) by domain, with the domain names of Table~\ref{tab:inventory} shortened. Humans are on
the 508 items (Table~\ref{tab:inventory} counts them by domain); models are on all pairs, the 508 items and the 5,942 pairs that humans
accepted (Pairs counts them). Column labels as in Table~\ref{tab:main}; BBL is BabyBabelLM 1M.}
\label{tab:domains}
\end{table*}

The second source is the Romanian part of BabyBabelLM \citep{jumelet-etal-2026-babybabellm}, which combines child-directed speech from CHILDES \citep{macwhinney-2000-childes,avram-2001-early,avram_childes}, children's books and OpenSubtitles \citep{lison2016opensubtitles}; in our training pool they make up 37.3\%, 29.3\% and 33.4\% of the words. We split each source 85\%/15\% by document, so that no document is divided between the two parts; for the captions, the split is stratified by age group.



GPT-2 follows the small-model recipe of \citet{feng2026babyscale}, and GPT-BERT \citep{charpentier-samuel-2024-bert} uses the BabyBabelLM implementation with equal shares of causal and masked objectives. Each condition has five runs with the training seeds 42, 0, 123, 321, and 9000, which also set the document order. From the held-out part we take 80,000 words of each source as one validation set of 160,000 words. Every model uses this same set, only to choose the checkpoint with the lowest loss, so that models of the same architecture differ only in their training data, as in \citet{hu-etal-2020-systematic}. Each comparison between training sets that we report holds constant the architecture, tokenizer, punctuation, and recipe. 

A model chooses the sentence with the higher  SLOR score \citep{pauls-klein-2012-large,lau-etal-2017}, computed as the log probability of the sentence minus its unigram log probability, divided by its number of tokens. For RoGPT2, GPT-2, GPT-BERT, EuroLLM and Salamandra, the log probability sums the token log probabilities given a beginning-of-text token, with GPT-BERT in causal mode. For RoBERT it is the word-level pseudo-log-likelihood of \citet{kauf-ivanova-2023-better}. 

Unigram probabilities use add-one smoothing over each model's tokens. For the models trained from scratch, we compute them using the training data. For the pretrained models, we estimate the frequencies from text similar to their training data \cite{lau-etal-2020-furiously}: Romanian Wikipedia and a 5\% sample of the Romanian part of CC-100 \citep{conneau-etal-2020-unsupervised,wenzek-etal-2020-ccnet}.

We define a model's accuracy on an item as the share of its runs that choose the target, and its score margin as the target's score minus the alternative's, averaged over runs. 

We compute 95\% intervals from 10,000 bootstrap samples \citep{efron-1979-bootstrap,efron-tibshirani-1993}. Each sample draws pairs at random, with replacement, within each phenomenon. For trained models it also draws among their five runs, and for people among the participants of each experiment. A pair that appears twice is drawn with both its presentations. Our procedure follows the Multi-Bootstrap of \citet{sellam-etal-2022-multiberts}, which also draws both test items and training runs. To compare two conditions, we compute their difference on the same draw of pairs and match runs trained with the same seed.

\section{Results}


Mean human item accuracy is 96\% and the two pretrained Romanian models reach 90.4\% and 92.3\% (Table~\ref{tab:main}). The human-minus-model difference is 5.7 points [3.4, 8.0] for RoGPT2 
and 3.7 points [1.7, 5.8] for RoBERT. RoGPT2 and RoBERT choose the same sentence as participants in an average of 88.1\% and 89.8\% of the 26,358 answers, while two participants choose the same sentence in 93.1\% of cases (Table~\ref{tab:main}).
The pretrained models score 94.7\% and 98.3\% on MultiBLiMP, so BLiMP-RO is the harder benchmark for them. Table~\ref{tab:main} reveals that, on the 508 items, GPT-BERT consistently outperforms GPT-2 across all limited-data training conditions, achieving 70--77\% accuracy compared with 65--73\% for GPT-2. Furthermore, increasing the caption training corpus from approximately 1M to 10M words improves BLiMP-RO accuracy by 7--8 percentage points for neural models. At matched training sizes, BabyBabelLM and child-targeted video captions yield comparable performance, with differences of about one percentage point for GPT-2 and GPT-BERT. Romanian-specific pretrained models achieve the highest BLiMP-RO accuracy (90--92\%), outperforming the multilingual models (82\%). This difference is considerably smaller on MultiBLiMP, where all four pretrained models achieve 94--98\% accuracy. 



\subsection{Domains}
\label{sec:domains}

The 508 items and the 5,942 human-validated pairs yield similar results: they rank the 19 models of Table~\ref{tab:main} almost identically, and for each small model, the two accuracies differ by at most 3.9 points. From here on, we report models on all 6,435 pairs together.

People are above 90\% in every categorical domain (Table~\ref{tab:domains}; by phenomenon, Figure~\ref{fig:phenomena} and Table~\ref{tab:phenomena} in Appendix~\ref{app:results}). They find DOM the hardest, at 90.1\%, because of the animacy items, and binding the next hardest, at 91.5\%. 

For models trained on 10M caption words, mood is the strongest domain (81.4--82.6\%), whereas binding remains challenging (58.5--59.6\%). Increasing the training data from 1M to 10M words improves performance across all domains for GPT-2, with particularly large gains in negation (+17.8 percentage points) and null subjects (+16.7 points). GPT-BERT exhibits a similar but non-uniform pattern, with its largest gains in nominal constructions (+13.6 points) and negation (+10.1 points). At matched training sizes, BabyBabelLM and caption data yield different domain-level profiles despite their comparable aggregate accuracy: BabyBabelLM generally favors clitics and word order, whereas captions yield better performance on null subjects. Pretrained models achieve substantially higher accuracy across domains, although binding remains among their weakest areas (86.2--88.6\%).

In seventeen categorical items in the agreement and binding domains, the target has a plural emphatic reflexive such as \emph{înșiși} `themselves' and the alternative a singular one; people average 82.6\% on them, and two of them are missed by both pretrained Romanian models.
\begin{figure}[!htb]
\centering
\includegraphics[width=\columnwidth]{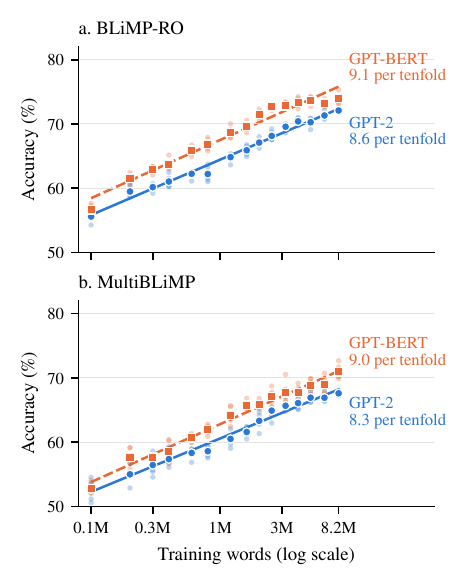}
\caption{Accuracy under SLOR of caption models against training words, on (a) BLiMP-RO,
all 6,435 pairs: the 493 items and the 5,942 pairs that humans accepted, and (b) MultiBLiMP. The 15 nested caption sets have 0.1, 0.2, 0.3, 0.4, 0.6, 0.8, 1.2, 1.6, 2, 2.5, 3.2, 4, 5,
6.4 and 8.2 million training words, without validation text; the last is the whole caption pool (captions 10M in
Table~\ref{tab:main}). All models use the same validation set of 160,000 words, half captions and half
BabyBabelLM, only to choose the checkpoint.
Small points are runs, large points their
means, lines ordinary least-squares fits of the means, and labels their gain in points per
tenfold increase.}
\label{fig:curve}
\end{figure}

Increasing the caption training data from 1M to 10M words improves BLiMP-RO accuracy by 9.9 percentage points [8.6, 11.2] for GPT-2 and 7.1 points [6.0, 8.1] for GPT-BERT. Because all models are trained for 20 epochs, larger datasets also entail more optimization steps. To disentangle these effects, we train the 1M models for the same number of updates as their 10M counterparts. The larger datasets retain an advantage of 10.1 points [8.6, 11.5] for GPT-2 and 6.7 points [5.6, 7.9] for GPT-BERT, indicating that the gains are not explained solely by additional optimization steps.

At 1M words, BabyBabelLM outperforms children's video captions by 5.4 points [4.0, 6.6] for GPT-2 and 3.9 points [2.7, 5.1] for GPT-BERT. To examine the contribution of individual data sources, we train both architectures on matched subsets of 238,600 words, using five runs per condition (Appendix \ref{app:results}). Children's books and subtitles yield approximately 4 points higher accuracy than captions, whereas CHILDES performs substantially worse. Lexical coverage may partly explain these differences: the 1M BabyBabelLM corpus contains 65.9\% of the 11,169 distinct words in BLiMP-RO, compared with 58.7\% for captions. By contrast, the targeted age group has a smaller effect. Models trained on 5M words of captions targeting ages 0--4 versus 5--8 differ by 2.1 points [0.9, 3.2] for GPT-2 and only 0.2 points [$-$1.0, 1.4] for GPT-BERT.

Figure~\ref{fig:curve} examines grammatical accuracy across 15 nested caption datasets ranging from 0.1M to 8.2M words. Accuracy increases approximately linearly with the logarithm of training-set size, with gains of 8.6 points [8.0, 9.2] per tenfold increase for GPT-2 and 9.1 points [8.4, 9.7] for GPT-BERT. Similar trends emerge on MultiBLiMP, with gains of 8.3 [7.1, 9.5] and 9.0 points [7.9, 10.2], respectively. The differences between benchmark slopes are small: 0.3 points [$-$1.0, 1.6] for GPT-2 and 0.0 points [$-$1.2, 1.3] for GPT-BERT, suggesting comparable scaling behavior across the two evaluation sets.

\section{Conclusion}

We introduced BLiMP-RO, a human-validated benchmark for evaluating grammatical knowledge in Romanian language models, covering 50 linguistic phenomena across ten grammatical domains. By combining human-constructed minimal pairs with a larger, semi-automatically generated and validated dataset, BLiMP-RO extends existing Romanian evaluation resources beyond agreement-focused phenomena. Our experiments show that pretrained Romanian models achieve high overall accuracy but exhibit substantial variation across grammatical domains, with persistent difficulties in constructions involving binding and syntactic extraction.

We further investigated how grammatical knowledge emerges under limited linguistic input by training Romanian GPT-2 and GPT-BERT models on BabyBabelLM and a newly collected corpus of child-targeted video captions. Grammatical accuracy increases approximately linearly with the logarithm of training-data size, with improvements persisting when the number of training updates is controlled. At matched data sizes, BabyBabelLM consistently outperforms video captions, while comparisons across its constituent sources suggest that children's books and subtitles provide useful training signals.

Our findings establish BLiMP-RO as a resource for investigating grammatical generalization in Romanian and provide evidence that relatively small language models can acquire measurable grammatical knowledge from developmentally motivated data. They also motivate further research into which properties of training input support the acquisition of specific grammatical constructions and how closely model behavior aligns with human linguistic judgments.

\section*{Ethics and Release}
Participants gave informed consent online and could withdraw. Responses were recorded under participant numbers rather than names. We will release the 528 items with their exact text and option order, keys, exclusions and their reasons, distinct-pair identifiers, pair types, item-level human accuracies, the model scores per item and run, and the analysis code. Because the caption text belongs to the video creators and YouTube's terms restrict its redistribution, we publicly release the video identifiers, metadata and hashes needed to rebuild the corpus, and share the caption text with researchers on request for research use only.

\section*{Limitations}
Human acceptability ratings were collected through forced-choice comparisons, which establish relative preferences but do not directly measure the acceptability of individual sentences. At the same time, the extended benchmark was human-validated using a custom-built interface. Therefore, the accuracy scores of models and humans are not directly comparable.

Our experiments establish associations between training-data properties and grammatical accuracy but do not identify the mechanisms underlying grammatical learning. Although we control for the number of optimization updates when comparing training-set sizes, other factors, including model architecture, tokenization, and training objectives, may influence performance.

\bibliography{tacl2021}
\bibliographystyle{acl_natbib}

\appendix
\renewcommand{\topfraction}{0.8}

\section{Study and Model Details}
\label{app:details}

The BabyBabelLM text is the Romanian 1M-word tier, \texttt{BabyLM-community/babylm-ron} at revision \texttt{caf295e}. We use the base RoGPT2 and RoBERT of ReaderBench, with 124.4M and 115.1M parameters at Hugging Face revisions a6584ccf and 9d0f3e69, and  EuroLLM-9B-2512 at revision 3f5bacd0 and Salamandra-7B at revision 0f5fa8e7. In causal scoring we prefix a space and the beginning-of-text token, or the end-of-text token if the tokenizer lacks one, and score every sentence token, with no end token. In tokenwise pseudo-log-likelihood scoring we mask each non-special token once, and in the word-level variant also the later tokens of its word, with the classification and separator tokens as context. The GPT-2 model has 12 layers, a 30,000-entry tokenizer and 108.9M parameters and trains for 20 epochs with Adam \citep{kingma-ba-2015-adam} at $10^{-4}$, batch size 8 and context length 1,024. GPT-BERT base has a 16,384-entry tokenizer and trains for 20 epochs with LAMB \citep{you-etal-2020-lamb} at $5{\times}10^{-3}$, a global batch of 256 and context length 128. Its BabyBabelLM implementation predicts each masked token at its own position, unlike the next-position prediction of the original GPT-BERT, and masks a fixed 30\% of tokens. With 1M training words, the checkpoint with the lowest validation loss comes from epochs 9 to 11 for GPT-2 trained on captions and 8 to 9 on BabyBabelLM, and from about epoch 10 to 15 for GPT-BERT trained on captions and about epoch 10 on BabyBabelLM. The 1M and 10M models of an architecture share one tokenizer, trained on 400,000 words of each source, and validation sets of 80,000 held-out words per source. Test sentences for the trained models receive the normalization of their training text: Unicode NFC, comma-below \emph{ș} and \emph{ț}, and single spaces. The unigram reference text for the pretrained models is configuration 20231101.ro of \texttt{wikimedia/wikipedia} at revision b04c8d1c and every 20th document of the Romanian CC-100 file, starting with the first, 1,510,524 of 30,210,462 documents.

We removed exact duplicates, near duplicates and short clips that also appear inside longer compilations. Near duplicates were found with MinHash \citep{broder1997minhash} over five-word sequences, using the datasketch library with 128 permutations, and confirmed when two documents share at least 72\% of these sequences or one contains at least 80\% of the other. A clip counts as part of a compilation when at least 80\% of its five-word sequences appear in the same order in the compilation. When the same content appeared in several documents, we kept one copy. We also removed 544 transcripts that contain a word from a fixed list of 69 profanities, and transcripts that are corrupted, not in Romanian or too short. No channel contributes more than 1M of the 5M words of an age group: we added the documents in a fixed random order and skipped any that would take its channel above this limit.

\paragraph{Caption audit.} The 40 windows come from 40 videos of 19 channels, 39.7 minutes in total, with 16 windows of automatic captions and 4 of creator captions for each age group, drawn with seed 42. The reference transcription was made from the audio with the captions hidden. Before the comparison, both texts are lowercased and stripped of punctuation, hyphenated forms are split, and digits are written as words. Sounds such as laughter and onomatopoeia are not counted, nor are proper names that the caption spells differently (29 of 103 names); a name that the caption leaves out still counts as an error. When both texts have a word or line but one repeats it more often, the extra repetitions are not counted as errors; a word or line missing from one text is. Short and full forms of the same word, such as \emph{îmi} and \emph{mi}, \emph{e} and \emph{i}, or \emph{este} and \emph{e}, are not counted as errors. Creator captions are rare in the corpus but make up a fifth of the sample, so the overall rate weights each window by how much of the corpus its group represents. The interval comes from 10,000 bootstrap samples of the windows. By group, the word error rate is 2.6\% for automatic captions of ages 0--4, 5.0\% for creator captions of ages 0--4, 5.7\% for automatic captions of ages 5--8 and 3.6\% for creator captions of ages 5--8. Without weights, it is 2.7\% in the 28 windows with speech, 7.8\% in the 9 with singing and 8.1\% in the 3 with both. In the reference transcription, 864 words carry agreement: finite verbs, auxiliaries, adjectives and determiners. The captions get 34 of them wrong, and in 9 of these the caption has the right word with a different person, number or gender, as in \emph{pot} `I can' for \emph{poți} `you can'.

\section{Other Scores}
\label{app:scores}

GPT-BERT also predicts masked tokens, and following \citet{feng2026babyscale} we read it in two more ways: each token masked in turn and predicted from the whole sentence, or with the causal and masked logits of each token summed before the softmax.

\begin{table}[ht]
\centering
\begin{tabular}{@{}lrrrr@{}}
\toprule
& Sum & Token & Word & SLOR \\
\midrule
\multicolumn{5}{@{}l}{\emph{GPT-2, left to right}} \\
\quad captions 1M & 53.0 & 64.9 & 55.1 & 64.6 \\
\quad BBL 1M & 55.6 & 65.8 & 55.8 & 65.2 \\
\quad captions 10M & 63.2 & 73.4 & 61.8 & 72.5 \\
\addlinespace[2pt]
\multicolumn{5}{@{}l}{\emph{GPT-BERT, left to right}} \\
\quad captions 1M & 57.9 & 69.7 & 57.3 & 70.1 \\
\quad BBL 1M & 62.0 & 70.8 & 61.6 & 71.1 \\
\quad captions 10M & 67.3 & 76.9 & 65.8 & 77.5 \\
\addlinespace[2pt]
\multicolumn{5}{@{}l}{\emph{GPT-BERT, bidirectional}} \\
\quad captions 1M & 59.8 & 72.2 & 59.9 & 71.7 \\
\quad BBL 1M & 66.0 & 72.6 & 66.4 & 73.0 \\
\quad captions 10M & 79.1 & 78.1 & 75.4 & 79.4 \\
\addlinespace[2pt]
\multicolumn{5}{@{}l}{\emph{GPT-BERT, fused}} \\
\quad captions 1M & 63.7 & 72.4 & 62.8 & 72.6 \\
\quad BBL 1M & 70.1 & 74.8 & 68.5 & 75.9 \\
\quad captions 10M & 78.9 & 79.7 & 75.4 & 80.2 \\
\addlinespace[2pt]
\multicolumn{5}{@{}l}{\emph{Pretrained, left to right}} \\
\quad RoGPT2 & 93.9 & 87.2 & 87.2 & 90.4 \\
\addlinespace[2pt]
\multicolumn{5}{@{}l}{\emph{Pretrained, bidirectional}} \\
\quad RoBERT & 92.5 & 90.7 & 91.9 & 92.3 \\
\bottomrule
\end{tabular}
\caption{Accuracy (\%) on the 508 items by how a model reads a sentence and how the reading is scored.
\emph{Left to right} predicts each token from the tokens before it; \emph{bidirectional} masks each token
in turn and predicts it from the rest of the sentence (pseudo-log-likelihood); \emph{fused} adds the two
readings' logits before the softmax, and only GPT-BERT is trained for both. Sum adds the log probabilities
of the sentence's tokens, Token and Word divide it by the number of tokens or words, and SLOR, the main
score of Table~\ref{tab:main}, subtracts the log probability of the same tokens under their frequencies
alone and divides by the number of tokens (Section~\ref{sec:models}); for RoBERT, a masked
token also hides the later tokens of its word in SLOR.}
\label{tab:scores}
\end{table}

\section{Additional Results}
\label{app:results}

\begin{table}[H]
\centering
\setlength{\tabcolsep}{3pt}
\begin{tabular}{@{}lrrrrrr@{}}
\toprule
& \multicolumn{4}{c}{Caption words} & & \\
\cmidrule(lr){2-5}
& 0.1M & 0.8M & 2M & 5M & Slope & $\Delta$BBL \\
\midrule
\multicolumn{7}{@{}l}{\emph{GPT-2}} \\
\quad order 1 & 55.6 & 62.2 & 67.1 & 70.3 & 8.8 & $-$5.4 \\
\quad order 2 & 55.0 & 63.0 & 66.1 & 70.6 & 9.0 & $-$4.6 \\
\quad order 3 & 56.6 & 63.5 & 66.8 & 70.0 & 8.0 & $-$4.1 \\
\addlinespace[2pt]
\multicolumn{7}{@{}l}{\emph{GPT-BERT}} \\
\quad order 1 & 56.6 & 66.9 & 71.5 & 73.6 & 9.4 & $-$3.9 \\
\quad order 2 & 56.4 & 66.2 & 71.1 & 72.8 & 9.3 & $-$4.6 \\
\quad order 3 & 56.8 & 67.2 & 71.0 & 73.5 & 9.2 & $-$3.5 \\
\bottomrule
\end{tabular}
\caption{Accuracy (\%) under SLOR on all 6,435 pairs for caption sets drawn from three random orders
of the caption videos (Section~\ref{sec:models}), five runs each; order 1 gives Figure~\ref{fig:curve}. With the whole pool, shared by all three, GPT-2
reaches 72.1\% and GPT-BERT 73.9\%. Slope: points gained per tenfold increase in words, from a
straight line through a row's five accuracies on a log scale. $\Delta$BBL: 0.8M captions minus BBL 1M.}
\label{tab:selections}
\end{table}

\begin{table}[H]
\centering
\begin{tabular}{@{}lrr@{}}
\toprule
Training text & GPT-2 & GPT-BERT \\
\midrule
Children's books & 63.8 & 66.4 \\
Subtitles & 64.0 & 65.9 \\
Captions & 59.7 & 62.4 \\
CHILDES & 53.1 & 50.5 \\
\bottomrule
\end{tabular}
\caption{Accuracy (\%) under SLOR on all 6,435 pairs of models trained on 238,600 words from each source, all the children's books in the BabyBabelLM training part, five runs each.}
\label{tab:sources}
\end{table}

\begin{table*}[p]
\centering
\footnotesize
\setlength{\tabcolsep}{4.3pt}
\renewcommand{\arraystretch}{0.95}
\begin{tabular}{@{}l rr rrr rrr rr@{}}
\toprule
& & & \multicolumn{3}{c}{GPT-2} & \multicolumn{3}{c}{GPT-BERT} & \multicolumn{2}{c}{Pretrained} \\
\cmidrule(lr){4-6}\cmidrule(lr){7-9}\cmidrule(l){10-11}
& Humans & Pairs & \begin{tabular}[b]{@{}r@{}}Cap.\\1M\end{tabular} & BBL & \begin{tabular}[b]{@{}r@{}}Cap.\\10M\end{tabular} & \begin{tabular}[b]{@{}r@{}}Cap.\\1M\end{tabular} & BBL & \begin{tabular}[b]{@{}r@{}}Cap.\\10M\end{tabular} & RoGPT2 & RoBERT \\
\midrule
\multicolumn{11}{@{}l}{\emph{Agreement}} \\
\quad subject-verb & 95.7 & 126 & 50.8 & 51.1 & 65.9 & 60.2 & 53.8 & 72.9 & 96 & 98.4 \\
\quad demonstrative & 96.2 & 129 & 60.8 & 64.8 & 74.9 & 66.8 & 67.6 & 74.9 & 98.4 & 100.0 \\
\quad adjective & 94.8 & 128 & 47.0 & 52.3 & 63.0 & 54.7 & 51.9 & 67.0 & 96.1 & 99.2 \\
\quad participial agreement & 98.8 & 130 & 51.5 & 50.6 & 54.6 & 54.0 & 50.6 & 58.8 & 92.3 & 99.2 \\
\quad adjectival article \emph{cel} & 96 & 129 & 57.7 & 52.9 & 75.7 & 66.5 & 65.0 & 79.7 & 99.2 & 100.0 \\
\quad anaphor agreement & 85.7 & 111 & 60.7 & 59.3 & 69.4 & 52.3 & 49.0 & 64.1 & 97.3 & 86.5 \\
\quad doubled object and clitic & 97.5 & 128 & 57.5 & 55.8 & 58.0 & 60.2 & 51.7 & 57.3 & 89.8 & 98.4 \\
\quad ordinal numeral & 99.0 & 130 & 52.3 & 56.6 & 61.1 & 53.8 & 55.4 & 64.6 & 95.4 & 97.7 \\
\multicolumn{11}{@{}l}{\emph{DOM}} \\
\quad \emph{pe} in object relatives & 96.4 & 130 & 67.8 & 80.3 & 82.2 & 92.0 & 94.5 & 94.8 & 100.0 & 95.4 \\
\quad \emph{pe} and clitic across noun types & 92.5 & 129 & 66.8 & 86.7 & 80.8 & 71.6 & 87.4 & 78.9 & 98.4 & 100.0 \\
\quad object marking across animacy classes & 81.4 & 130 & 30.9 & 30.6 & 38.6 & 40.3 & 30.0 & 46.3 & 83.8 & 92.3 \\
\multicolumn{11}{@{}l}{\emph{Clitics}} \\
\quad clitic doubling & 95.1 & 234 & 68.4 & 73.0 & 87.2 & 73.9 & 84.8 & 86.4 & 98.4 & 98.4 \\
\quad resumptive clitics & 97.1 & 125 & 74.2 & 78.2 & 85.8 & 85.1 & 86.6 & 90.9 & 99.2 & 98.4 \\
\quad clitic order, Person Case Constraint & 97.5 & 120 & 69.0 & 68.7 & 88.2 & 79.3 & 83.3 & 82.7 & 98.3 & 97.5 \\
\quad clitics with modal verbs & 96.7 & 126 & 85.7 & 86.0 & 87.3 & 91.7 & 91.3 & 91.0 & 96.8 & 90.5 \\
\quad clitics in negative imperatives & 97.6 & 122 & 47.0 & 81.5 & 29.5 & 58.5 & 67.2 & 37.4 & 66.4 & 60.7 \\
\multicolumn{11}{@{}l}{\emph{Binding}} \\
\quad case condition & 97.3 & 130 & 58.3 & 68.6 & 69.8 & 55.1 & 65.1 & 76.5 & 100.0 & 100.0 \\
\quad C-command condition & 82.7 & 112 & 50.5 & 50.9 & 46.8 & 46.3 & 50.9 & 46.6 & 84.8 & 76.8 \\
\quad locality of reflexives & 94.6 & 127 & 51.8 & 56.2 & 57.2 & 47.2 & 50.1 & 53.7 & 80.3 & 80.3 \\
\multicolumn{11}{@{}l}{\emph{Arguments}} \\
\quad transitivity & 99.1 & 128 & 53.6 & 60.3 & 62.0 & 60.5 & 63.0 & 62.3 & 80.5 & 95.3 \\
\quad passivizability & 98.7 & 125 & 45.1 & 49.3 & 62.2 & 57.1 & 54.4 & 66.7 & 96 & 100.0 \\
\quad inherent reflexive verbs & 95.0 & 130 & 69.2 & 77.2 & 79.5 & 70.8 & 84.6 & 80.6 & 100.0 & 100.0 \\
\multicolumn{11}{@{}l}{\emph{Word order}} \\
\quad relativization & 98.7 & 127 & 51.8 & 77.5 & 65.0 & 53.4 & 77.5 & 76.7 & 99.2 & 99.2 \\
\quad demonstrative placement & 94.3 & 129 & 78.3 & 82.3 & 88.4 & 78.1 & 80.3 & 84.5 & 97.7 & 100.0 \\
\quad topicalization & 97.7 & 130 & 76.2 & 79.1 & 88.0 & 77.5 & 91.5 & 85.2 & 100.0 & 99.2 \\
\quad definite article placement & 98.5 & 130 & 57.1 & 58.5 & 60.2 & 58.3 & 59.5 & 60.6 & 93.1 & 97.7 \\
\quad indefinite article placement & 99.0 & 130 & 95.2 & 98.6 & 92.9 & 95.1 & 99.5 & 96.5 & 100.0 & 100.0 \\
\quad preverbal depictives & 95.2 & 127 & 76.9 & 86.9 & 86.3 & 76.5 & 86.1 & 86.8 & 99.2 & 99.2 \\
\quad auxiliary and verb order & 94.8 & 130 & 98.6 & 99.5 & 99.5 & 98.9 & 100.0 & 99.4 & 100.0 & 100.0 \\
\quad adpositional phrases & 100.0 & 125 & 76.6 & 79.4 & 91.0 & 82.9 & 87.2 & 93.3 & 100.0 & 99.2 \\
\quad extraction from complex noun phrases & 98.0 & 126 & 59.0 & 58.4 & 56.5 & 58.4 & 70.5 & 56.3 & 85.7 & 96 \\
\quad extraction from adjuncts & 99.4 & 125 & 23.7 & 25.0 & 51.7 & 37.1 & 36.2 & 49.0 & 52.8 & 76.8 \\
\quad extraction from coordinations & 99.2 & 121 & 61.8 & 61.8 & 58.7 & 60.5 & 68.6 & 72.4 & 95.9 & 93.4 \\
\quad extraction from embedded questions & 98.2 & 130 & 37.1 & 43.5 & 59.4 & 44.8 & 58.0 & 49.1 & 75.4 & 79.2 \\
\quad extraction from clausal subjects & 99.2 & 127 & 82.0 & 88.0 & 90.7 & 84.3 & 88.3 & 83.5 & 95.3 & 99.2 \\
\multicolumn{11}{@{}l}{\emph{Mood}} \\
\quad subjunctive-selecting predicates & 96.9 & 130 & 78.9 & 92.9 & 88.9 & 86.2 & 94.9 & 92.0 & 98.5 & 100.0 \\
\quad indicative-selecting predicates & 98.3 & 130 & 72.6 & 62.6 & 65.4 & 66.0 & 58.8 & 63.8 & 90.8 & 89.2 \\
\quad subjunctive and infinitive* & 96.4 & 130 & 78.0 & 78.9 & 92.8 & 91.8 & 87.4 & 94.6 & 100.0 & 100.0 \\
\quad conditional & 96.8 & 130 & 70.3 & 83.1 & 74.5 & 76.5 & 80.2 & 73.7 & 96.2 & 98.5 \\
\quad future & 96.4 & 129 & 78.8 & 76.7 & 91.3 & 79.1 & 77.5 & 82.8 & 98.4 & 98.4 \\
\multicolumn{11}{@{}l}{\emph{Negation}} \\
\quad negative concord & 97.9 & 129 & 49.1 & 64.3 & 67.9 & 60.9 & 65.9 & 78.3 & 96.9 & 100.0 \\
\quad negative additive \emph{nici} & 98.1 & 126 & 56.8 & 72.2 & 86.3 & 72.9 & 84.3 & 86.7 & 94.4 & 96 \\
\quad negative imperatives & 88.0 & 108 & 56.3 & 62.0 & 59.3 & 62.4 & 58.5 & 59.6 & 70.4 & 72.2 \\
\multicolumn{11}{@{}l}{\emph{Nominal}} \\
\quad gender and number & 95.4 & 128 & 65.3 & 64.1 & 69.4 & 66.9 & 61.1 & 68.8 & 96.9 & 100.0 \\
\quad plural forms & 96.3 & 128 & 51.9 & 52.2 & 67.5 & 62.7 & 64.4 & 75.5 & 98.4 & 98.4 \\
\quad nominalization & 93.5 & 131 & 55.1 & 57.9 & 69.4 & 55.9 & 56.2 & 74.1 & 95.6 & 96.3 \\
\quad genitive and dative & 98.4 & 130 & 67.5 & 74.9 & 76.5 & 67.1 & 69.1 & 75.5 & 97.7 & 100.0 \\
\quad cardinal numerals with \emph{de} & 97.6 & 126 & 40.8 & 67.8 & 56.5 & 25.2 & 66.7 & 59.7 & 100.0 & 96.8 \\
\quad vocative & 98.4 & 122 & 70.2 & 83.8 & 75.4 & 75.1 & 84.6 & 80.7 & 90.2 & 74.6 \\
\multicolumn{11}{@{}l}{\emph{Null subjects}} \\
\quad null against overt subject* & 97.8 & 122 & 57.7 & 47.0 & 74.4 & 79.8 & 69.3 & 84.3 & 99.2 & 100.0 \\
\bottomrule
\end{tabular}
\caption{Accuracy (\%) by phenomenon, columns as in Table~\ref{tab:domains}. *Preference phenomena.}
\label{tab:phenomena}
\end{table*}

\end{document}